%% file: main.tex
\documentclass[11pt,twocolumn]{article}
\usepackage[a4paper,top=0.75in,bottom=1in,left=0.75in,right=0.75in,columnsep=0.2in]{geometry}
\usepackage[T1]{fontenc}
\usepackage{lmodern}
\usepackage{times} 
\usepackage{setspace}
\usepackage{amsmath, amssymb, graphicx, url, booktabs}
\usepackage{titlesec}
\usepackage{parskip} 
\usepackage[sorting=none]{biblatex}
\usepackage{fancyhdr}
\usepackage{hyperref}
\usepackage{lastpage}

\titleformat{\section}
  {\normalfont\fontsize{12}{14}\bfseries\centering\scshape}{\thesection.}{1em}{}
\titlespacing{\section}{0pt}{8pt}{8pt}

\titleformat{\subsection}
  {\normalfont\fontsize{12}{14}\itshape}{\thesubsection.}{1em}{}
\titlespacing{\subsection}{0pt}{6pt}{6pt}

\begin{document}

\thispagestyle{empty} 
\twocolumn[
\begin{center}
{\fontsize{24}{28}\selectfont Application of Deep Learning in Biological Data Compression} 

\vspace{12pt}

{\fontsize{12}{14}\selectfont Chunyu Zou}

{\fontsize{12}{14}\selectfont Department of Electrical and Electronic Engineering, The University of Hong Kong\\ Email: u3594662@connect.hku.hk}
\end{center}
\vspace{24pt}
]

\setcounter{page}{1}

\noindent\textbf{\textit{Abstract -}} \textbf{\input{Abstract}}

\vspace{12pt}

\noindent\textit{Index Terms -} Biological data compression, Data representation, Implicit neural representation, Neural Network

\input{Introduction}
\input{Methodology}
\input{Discussion}
\input{Conclusion}

\printbibliography

\end{document}

%% file: Abstract.tex
\noindent Cryogenic electron microscopy (Cryo-EM) has become an essential tool for capturing high-resolution biological structures. Despite its advantage in visualizations, the large storage size of Cryo-EM data file poses significant challenges for researchers and educators. This paper investigates the application of deep learning, specifically implicit neural representation (INR), to compress Cryo-EM biological data. The proposed approach first extracts the binary map of each file according to the density threshold. The density map is highly repetitive, ehich can be effectively compressed by GZIP. The neural network then trains to encode spatial density information, allowing the storage of network parameters and learnable latent vectors. To improve reconstruction accuracy, I further incorporate the positional encoding to enhance spatial representation and a weighted Mean Squared Error (MSE) loss function to balance density distribution variations. Using this approach, my aim is to provide a practical and efficient biological data compression solution that can be used for educational and research purpose, while maintaining a reasonable compression ratio and reconstruction quality from file to file.

%% file: Introduction.tex
\section{Introduction}
\setlength{\parindent}{0pt}
Artificial intelligence (AI) has undergone remarkable advancements over the past few decades, transitioning from rudimentary rule-based systems to highly sophisticated models driven by deep learning techniques. Deep learning, a specialized subset of machine learning, leverages artificial neural networks with multiple interconnected layers to extract and process intricate patterns from vast and complex datasets \cite{lecun2015deep}. This paradigm shift has been fueled by the exponential growth in computational power, the availability of large-scale annotated datasets, and breakthroughs in optimization algorithms such as stochastic gradient descent \cite{goodfellow2016deep}. The success of deep learning has catalyzed transformative applications across diverse domains, including image recognition, speech processing, natural language understanding, and, more recently, the analysis of biological data \cite{min2017deep}. These advancements have not only enhanced the accuracy of predictive models but also expanded the scope of problems that AI can address, making it an indispensable tool in modern scientific research.

One of the most promising areas where deep learning is making a profound impact is in the field of structural biology, particularly through its application to Cryogenic Electron Microscopy (Cryo-EM). Cryo-EM is a cutting-edge imaging technique that has emerged as a cornerstone of modern structural biology, enabling scientists to capture high-resolution, three-dimensional structures of biological macromolecules in their native, hydrated states \cite{cheng2005single}. Unlike traditional methods such as X-ray crystallography, which require crystallization of samples—a process that can be challenging or impossible for certain molecules—Cryo-EM offers a more versatile approach by imaging samples preserved in a thin layer of vitreous ice \cite{frank2006three}. This preservation is achieved by rapidly freezing the biological specimens at cryogenic temperatures, typically using liquid ethane or nitrogen, to prevent the formation of ice crystals that could damage the molecular structure \cite{kuhlbrandt2014cryo}. Once frozen, the samples are subjected to a beam of electrons in a transmission electron microscope, producing a series of two-dimensional projection images. These projections are then computationally processed using advanced reconstruction algorithms, such as single-particle analysis, to generate a detailed three-dimensional volumetric representation of the molecule \cite{glaeser2016how}. The resulting data, often stored in standardized file formats such as MRC or MAP, encapsulates volumetric density values that reflect the spatial distribution of electrons within the sample. Visualization tools like ChimeraX or UCSF Chimera allow researchers to explore these density maps, adjusting thresholds to highlight regions of interest, such as high-density areas corresponding to well-defined molecular features \cite{goddard2018ucsf}.

The advent of Cryo-EM has revolutionized structural biology by providing unprecedented insights into the architecture of complex biomolecules, including proteins, viruses, and cellular machinery, which were previously beyond the reach of traditional imaging techniques \cite{kuhlbrandt2014cryo}. For instance, Cryo-EM has enabled the determination of atomic-resolution structures of membrane proteins and large macromolecular complexes, shedding light on their functional mechanisms and aiding drug design efforts \cite{bai2015cryo}. However, despite its transformative potential, Cryo-EM comes with significant practical challenges, primarily due to the sheer volume of data it generates. A single Cryo-EM experiment can produce files ranging from hundreds of megabytes to several gigabytes in size, depending on the resolution and the number of projections collected \cite{kuhlbrandt2014cryo}. For a comprehensive study, researchers may generate dozens or even hundreds of such files, leading to datasets that easily accumulate into tens or hundreds of gigabytes. This data deluge poses substantial hurdles in terms of storage, computational processing, and data sharing, particularly in resource-constrained environments such as academic institutions or educational settings where multiple users require simultaneous access \cite{nogales2016development}.
Moreover, the nature of Cryo-EM data introduces additional complexities. When visualizing a Cryo-EM density map, researchers typically apply a density threshold to isolate regions of high electron density, which correspond to well-resolved molecular structures \cite{frank2006three}. Lower-density regions, while sometimes informative, often contain noise or artifacts that obscure meaningful interpretation. The raw Cryo-EM data rarely exhibits negative density values; however, post-processing techniques—such as Gaussian filtering or B-factor sharpening—can alter the density distribution, pushing near-zero values into the negative range \cite{glaeser2016how}. These negative densities are generally considered artifacts and contribute little to structural understanding, yet they remain embedded in the data, inflating file sizes with redundant or irrelevant information. This inefficiency exacerbates the storage and computational burden, making it difficult to manage Cryo-EM datasets effectively. For example, a high-resolution map of a large protein complex might contain millions of voxels (three-dimensional pixels), many of which represent noisy or low-density regions that could be omitted without significant loss of scientific value \cite{cheng2015primer}. Consequently, the large file sizes and inherent redundancies hinder real-time visualization, data transfer, and collaborative research efforts, limiting the broader adoption of Cryo-EM in both research and educational contexts.

Deep learning has emerged as a powerful tool for addressing complex data challenges, including those posed by Cryo-EM. Its ability to model high-dimensional data and predict properties from raw inputs has made it a focal point of research across multiple disciplines \cite{GIRI2023102536}. In the context of Cryo-EM, deep learning has already demonstrated its utility in several key areas of data processing and analysis. For instance, convolutional neural networks (CNNs) have been employed to enhance the quality of Cryo-EM maps, as seen in tools like DeepEMhancer, which refines experimental maps by learning from paired datasets of raw and sharpened maps \cite{sanchez2021deepemhancer}. Similarly, super-resolution techniques have been developed to push the resolution of Cryo-EM maps beyond experimental limits, providing finer structural details \cite{subramaniya2021super}. Beyond map enhancement, deep learning has also been applied to particle picking—the process of identifying and extracting individual molecular images from noisy micrographs—with tools like DeepPicker automating and improving the accuracy of this labor-intensive step \cite{wang2016deeppickerdeeplearningapproach}. Perhaps the most celebrated example is AlphaFold, which leverages deep learning to predict protein structures from amino acid sequences with remarkable precision, complementing experimental techniques like Cryo-EM \cite{jumper2021highly}.

Despite these advancements, one area that remains underexplored is the application of deep learning to Cryo-EM data compression. Current storage practices, such as those employed by the Electron Microscopy Data Bank (EMDB), rely heavily on traditional compression methods like GZIP or store data in uncompressed formats \cite{lawson2011emdb}. While GZIP, based on algorithms like Huffman coding, can reduce file sizes by identifying and encoding repetitive patterns, it struggles with the noisy, non-repetitive nature of Cryo-EM density maps \cite{frank2006three}. The spatial complexity and variability in these datasets mean that traditional methods achieve only modest compression ratios, often insufficient to address the storage and transfer challenges faced by the Cryo-EM community. This gap in the literature presents an opportunity to explore innovative deep learning-based approaches that can better capture the underlying structure of Cryo-EM data and achieve higher compression efficiency.

This project proposes the use of Implicit Neural Representation (INR) as a novel deep learning-based method for compressing Cryo-EM data. Unlike traditional compression techniques that rely on explicit data encoding, INR represents the spatial density information of a Cryo-EM map as a continuous function parameterized by a neural network \cite{sitzmann2020implicit}. By training the network to map spatial coordinates to density values, INR encodes the data into a compact set of model parameters and learnable vectors, discarding the need to store the full voxel grid explicitly. This approach has the potential to achieve significantly higher compression ratios while preserving the fidelity of the reconstructed data, making it a promising solution for the storage and dissemination of Cryo-EM datasets.

To realize efficient Cryo-EM data compression, this project pursues two primary objectives:
\begin{enumerate}
    \item \textbf{Compression Ratio:} This metric quantifies the reduction in file size achieved by the INR method, defined as the ratio of the original file size to the compressed representation’s size. The goal is to maximize this ratio while ensuring that the compressed data retains sufficient quality for scientific analysis \cite{martel2021implicit}.
    \item \textbf{Reconstruction Quality:} This measures the accuracy with which the INR model reconstructs the original density map. Quality is assessed by comparing the reconstructed data to the original using quantitative metrics such as Mean Squared Error (MSE), Peak Signal-to-Noise Ratio (PSNR), and relative error at specific points of interest \cite{sitzmann2020implicit}.
\end{enumerate}

The successful implementation of this INR-based compression method could have far-reaching implications for the Cryo-EM community. By reducing file sizes without compromising data integrity, it would alleviate storage constraints, accelerate data sharing across research institutions, and enable real-time visualization on standard hardware \cite{nogales2016development}. This, in turn, could democratize access to Cryo-EM data, empowering researchers and educators with limited resources and potentially accelerating discoveries in structural biology, such as the elucidation of novel protein structures or the design of targeted therapeutics.

The remainder of this report is structured as follows: Section 2 details the realted work from this study. Section 3 gives the methodology, encompassing data preprocessing, INR network design, and training procedures. Section 4 presents preliminary results from applying the method to three Cryo-EM datasets, with quantitative evaluations of compression and reconstruction performance. Section 5 discusses implementation challenges and outlines a future work plan with specific milestones. Finally, Section 6 offers concluding remarks on the project’s progress and its potential contributions to the field.
\setlength{\parindent}{15pt} 

\section{Related Work}
\setlength{\parindent}{0pt}
The intersection of deep learning and Cryo-EM has been a fertile ground for research, with numerous studies exploring how neural networks can enhance various aspects of Cryo-EM data processing. This section reviews key prior works relevant to this project, focusing on deep learning applications in Cryo-EM analysis and compression, as well as foundational developments in implicit neural representations.

\subsection{Deep Learning in Cryo-EM Data Processing}
The application of deep learning to Cryo-EM has primarily targeted three areas: particle picking, map enhancement, and structural prediction. In particle picking, the identification of individual molecular particles within noisy micrographs is a critical and time-consuming step. Wang et al. \cite{wang2016deeppickerdeeplearningapproach} introduced DeepPicker, a convolutional neural network (CNN)-based tool that automates this process. Trained on labeled micrographs, DeepPicker achieves higher accuracy and efficiency than traditional heuristic methods, reducing manual effort and improving the quality of downstream reconstructions. Subsequent works, such as Topaz by Bepler et al. \cite{bepler2019topaz}, have further refined particle picking by incorporating unsupervised learning techniques to handle diverse datasets with minimal annotation.

In the domain of map enhancement, deep learning has been used to improve the interpretability of Cryo-EM density maps. DeepEMhancer, developed by Sanchez-Garcia et al. \cite{sanchez2021deepemhancer}, employs a CNN to post-process experimental maps, enhancing their sharpness and reducing noise. The model is trained on pairs of raw and high-quality maps, learning to mimic expert-level sharpening techniques. Similarly, Subramaniya et al. \cite{subramaniya2021super} proposed a super-resolution framework that leverages generative adversarial networks (GANs) to upscale low-resolution Cryo-EM maps, achieving resolutions beyond experimental limits. These advancements demonstrate the power of deep learning in refining Cryo-EM data but do not address the underlying issue of data size.

Perhaps the most prominent example of deep learning in structural biology is AlphaFold, developed by Jumper et al. \cite{jumper2021highly}. While not directly tied to Cryo-EM data processing, AlphaFold’s ability to predict protein structures from amino acid sequences with near-experimental accuracy has complemented Cryo-EM workflows. By providing high-confidence structural models, AlphaFold reduces the reliance on extensive experimental data collection, though it does not mitigate the storage demands of Cryo-EM datasets themselves \cite{jumper2021highly}.

\subsection{Data Compression in Cryo-EM}
Efforts to compress Cryo-EM data have largely relied on traditional methods. The Electron Microscopy Data Bank (EMDB), a primary repository for Cryo-EM maps, typically stores data either uncompressed or compressed using GZIP \cite{lawson2011emdb}. GZIP, based on the DEFLATE algorithm, combines Huffman coding and LZ77 to exploit redundancy in data \cite{deutsch1996gzip}. However, its effectiveness is limited in Cryo-EM contexts due to the noisy, non-repetitive nature of density maps. Studies have shown that GZIP achieves compression ratios of approximately 2:1 to 3:1 for Cryo-EM files, far below what is needed for efficient storage and transfer of large datasets \cite{kuhlbrandt2014cryo}. Alternative approaches, such as lossy compression techniques tailored to volumetric data (e.g., JPEG-like methods for 3D grids), have been proposed but are rarely adopted due to concerns over loss of scientific fidelity \cite{cheng2015primer}.

\subsection{Implicit Neural Representations}
Implicit Neural Representations (INRs) represent a paradigm shift in data encoding, moving away from discrete grids to continuous functions parameterized by neural networks. Introduced by Sitzmann et al. \cite{sitzmann2020implicit}, INRs have been applied to compress and represent various data types, including images, 3D shapes, and scenes. By training a neural network to map coordinates to signal values (e.g., pixel intensities or densities), INRs achieve compact representations that scale with model complexity rather than data resolution. In the context of scientific data, INRs have been explored for compressing volumetric datasets, such as those from medical imaging (e.g., CT scans), with promising results in balancing compression ratios and reconstruction quality \cite{martel2021implicit}. However, their application to Cryo-EM data remains largely uncharted, presenting a novel research direction for this project.

\subsection{Gap in the Literature}
While deep learning has advanced Cryo-EM data processing in areas like particle picking and map enhancement, and INRs have shown potential in other domains, there is a notable lack of research applying neural-based compression to Cryo-EM datasets. This project bridges this gap by adapting INR to the unique characteristics of Cryo-EM data, aiming to achieve higher compression ratios and maintain reconstruction fidelity compared to traditional methods \cite{sitzmann2020implicit}.
\setlength{\parindent}{15pt} 

%% file: Methodology.tex

\section{Method}

\setlength{\parindent}{0pt}

This paper presents a method to compress Cryogenic Electron Microscopy (Cryo-EM) data using Implicit Neural Representation (INR), a deep learning approach that encodes spatial data into compact neural network parameters. Cryo-EM, widely used in structural biology, generates large datasets that challenge storage and computational resources. The proposed INR method addresses these issues by transforming 3D coordinates from Cryo-EM files into a high-dimensional representation via positional encoding, concatenating them with a random identifier, and processing them in chunks to predict 1D density values. This approach achieves a balance between compression ratio and reconstruction fidelity, making Cryo-EM data more accessible for research and education. The network structure is depicted in Figure~\ref{fig:INR}.

This section is structured as follows: Subsection 2.1 provides an in-depth analysis of the compression challenges, Subsection 2.2 elaborates on the data preparation process with practical considerations, and Subsection 2.3 details the INR network structure, including positional encoding, identifier design, and the weighted loss function.

\subsection{Problem Scenario}

\begin{figure}[h]
    \centering
    \includegraphics[width=\linewidth]{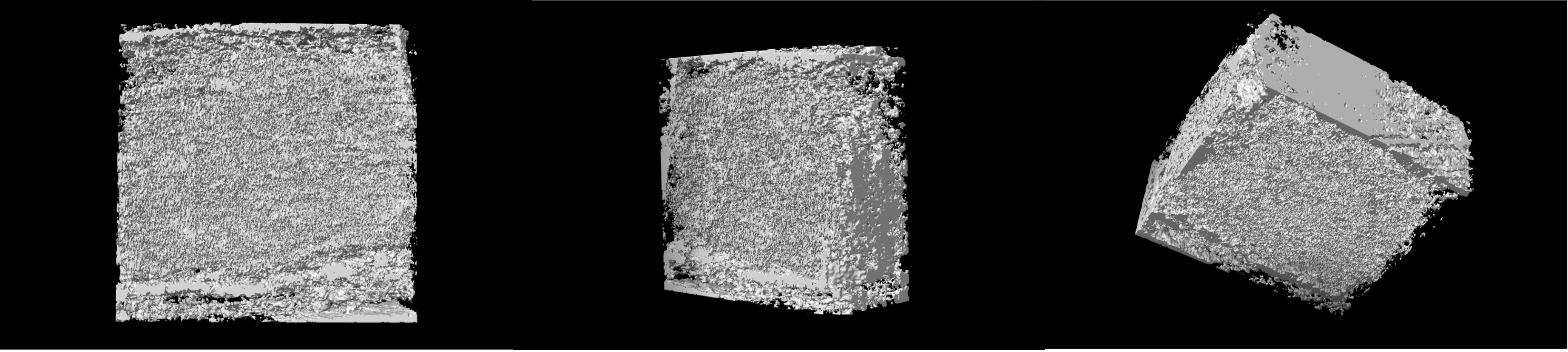}
    \caption{Density distribution of a sample Cryo-EM file, showing varying density values across a 3D volume. This complexity must be preserved during compression.}
    \label{fig:ds}
\end{figure}

The primary goal of Cryo-EM data compression is to minimize reconstruction error while achieving a high compression ratio, defined as the ratio of the original file size to the compressed size. Figure~\ref{fig:ds} illustrates the density distribution of a sample Cryo-EM file, revealing the intricate 3D spatial patterns that must be preserved.

Two significant challenges complicate this task:
\begin{enumerate}
    \item \textbf{Data Noise and Irregularity}: Cryo-EM data inherently contains noise from imaging processes and irregular point distributions due to molecular variability. For instance, a typical Cryo-EM map may include low-density noise from background scattering alongside high-density regions representing protein structures. Filtering this noise, as done in traditional preprocessing, risks losing subtle structural details critical for biological analysis, such as side-chain orientations. This necessitates a network design that can learn from noisy, irregular data without excessive preprocessing, preserving the full spectrum of spatial features.
    \item \textbf{Memory Constraints}: Each Cryo-EM file comprises approximately 100 million points, occupying ~414 MB in raw form. When combined with model parameters (e.g., weights and biases) and computational overhead (e.g., gradient calculations), the memory footprint can exceed typical GPU capacities, such as 12 GB on an NVIDIA GTX 1080 Ti or 24 GB on an RTX 3090. This is particularly problematic for end users—researchers or students with standard hardware—highlighting the need for a memory-efficient solution that scales to practical deployment scenarios.
\end{enumerate}

These challenges undermine conventional compression methods. Traditional models like vanilla autoencoders struggle with noisy data, as their latent representations often fail to capture irregular spatial relationships, leading to blurred reconstructions. Similarly, memory-intensive approaches that load entire datasets into GPU memory are impractical for large-scale Cryo-EM studies, where multiple files are processed simultaneously. For example, a dataset of 10 files (~4 GB total) exceeds even high-end GPU memory, causing crashes or requiring expensive hardware upgrades. The INR-based method overcomes these limitations by processing data in manageable chunks and leveraging neural encoding to tolerate irregularity, offering a scalable alternative to existing techniques.

\subsection{Data Preparation}

Effective compression begins with preprocessing raw Cryo-EM data to reduce redundancy and optimize memory usage. Cryo-EM files, typically stored in formats like MRC, contain 3D voxel grids with density values reflecting electron scattering intensity. Post-processing steps, such as sharpening to enhance map contrast, often introduce negative density values (e.g., -0.3 in normalized units). These negative values, while useful for visualization, are redundant for many research and educational purposes, as they often represent background noise rather than molecular structure. To address this, a density threshold hyperparameter is introduced, filtering out values below a user-defined limit (e.g., 0 in normalized units). This thresholding preserves meaningful density data while reducing the dataset size.

To track filtered points, binary occupancy maps are generated for each file. These maps assign a value of 1 to coordinates with density above the threshold and 0 otherwise, based on comparisons with corresponding density values. Early experiments explored using INR to predict occupancy directly, hypothesizing that a neural network could learn spatial occupancy patterns. However, analysis revealed that these maps exhibit high repetition—long sequences of 0s or 1s due to contiguous density regions—making them ideal for traditional compression tools like GZIP. GZIP achieves significant size reduction on these maps (e.g., from 100 MB to 10 MB for a typical file), complementing the INR’s role in compressing density data. This hybrid approach leverages the strengths of both methods: GZIP for repetitive binary data and INR for complex spatial density.

The filtered dataset, consisting of normalized 3D coordinates (e.g., [0, 1] range) and their corresponding density values, is stored in HDF5 format using the h5py Python library. HDF5, a hierarchical data format, supports efficient chunk-wise access, reducing GPU memory demands during training. For instance, instead of loading a 414 MB file entirely, the model accesses 100 tousand points at once, fitting within typical memory constraints. This preprocessing pipeline—thresholding, binary map generation, and HDF5 storage—ensures the data is both compact and accessible, setting the stage for neural compression.

\subsection{Network Structure}

\begin{figure}[h]
    \centering
    \includegraphics[width=\linewidth]{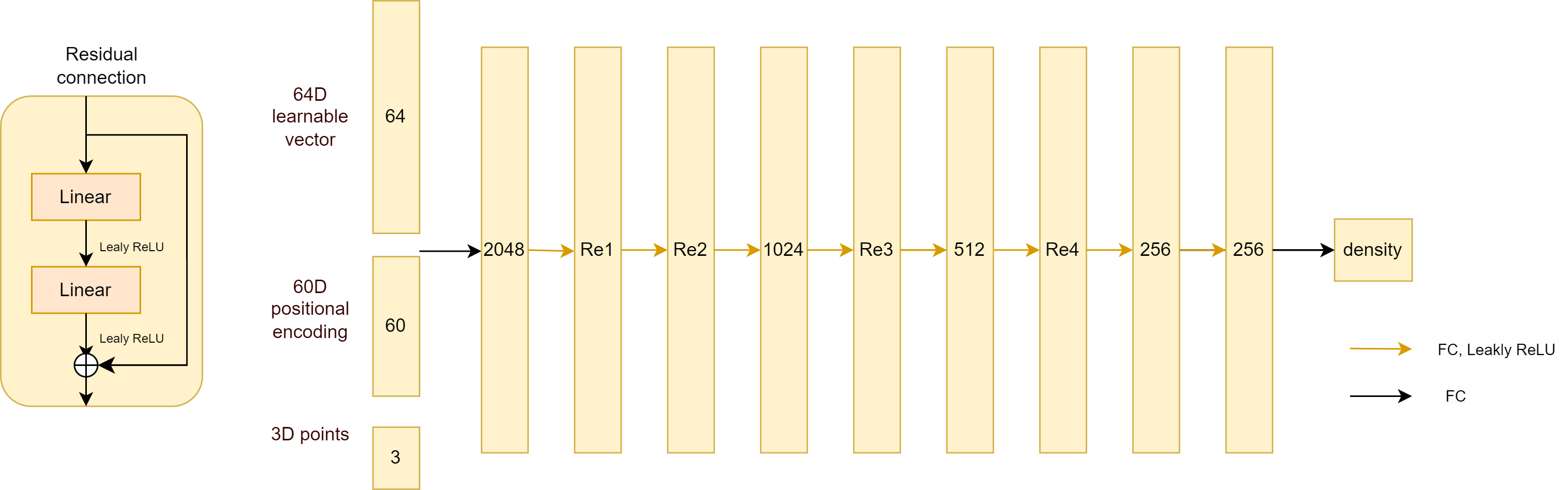}
    \caption{Network structure for the INR approach. 3D coordinates undergo positional encoding, are concatenated with a 64D random identifier, and pass through linear layers to predict density values. The network undergoes residual connection, resulting in better performance.}
    \label{fig:INR}
\end{figure}

The INR architecture, illustrated in Figure~\ref{fig:INR}, is designed to minimize reconstruction error while maximizing compression ratio, addressing the challenges outlined in Subsection 2.1.

The network processes HDF5-stored 3D coordinates and 1D density values in chunks to manage the ~414 MB file size. Chunking divides the dataset into smaller batches (e.g., 1 million points per chunk), enabling training on standard GPUs without memory overflow. Each chunk’s 3D coordinates are transformed via positional encoding, inspired by Neural Radiance Fields (NeRF)~\cite{mildenhall2020nerf}, from 3D to 63D (60D encoded + 3D original). The encoding function $\gamma(p)$ is defined as:

\begin{equation}
\begin{aligned}
    \gamma(p) = \big( &\sin(2^0 \pi p), \cos(2^0 \pi p), \ldots, \\
                   &\sin(2^9 \pi p), \cos(2^9 \pi p) \big),
\end{aligned}
\label{eq:enc}
\end{equation}

where $p$ is a 3D coordinate, and the frequency terms range from $2^0$ to $2^9$ (10 pairs, yielding 60 dimensions). This multi-frequency encoding maps low-dimensional coordinates into a higher-dimensional space, amplifying differences between adjacent points (e.g., from 0.01 to 0.1 in sine terms). This enhances the network’s ability to capture fine spatial details, critical for reconstructing complex Cryo-EM density maps.

Since multiple files are trained simultaneously, distinguishing identical coordinates across files is essential. Initial tests with an autoencoder produced overlapping latent vectors, reducing file-specific accuracy. To address this, a 64D Learnable vector is concatenated with the 63D encoded vector, forming a 127D input. The 64 size was chosen after experimentation: lower dimensions (e.g., 32D) risked collisions, while higher dimensions (e.g., 256) increased memory use without significant gain. Random numbers are generated once in the first epoch using a uniform distribution (e.g., [-1, 1]) and stored for reuse, ensuring consistency across training iterations.

The 127D input passes through a series of linear lay-
ers with Leaky ReLU activations (slope 0.01), pro-
gressively reducing dimensionality to a 1D density
output. The architecture comprises nine layers (e.g.,
127-2048-Re1-Re2-1024-Re3-512-Re4-256-1, where
Re refers to residual connection), balancing model ca-
pacity and compression efficiency. Standard Mean
Squared Error (MSE),

\begin{equation}
    \text{MSE}(y, \hat{y}) = (y - \hat{y})^2,
    \label{eq:mse}
\end{equation}

where $y$ is the ground truth and $\hat{y}$ is the prediction, was initially used but proved inadequate. Density values cluster heavily near zero (~70\% within [-0.01, 0.01], range -0.3 to 0.2), causing the model to prioritize low-value regions and underfit outliers. This resulted in a narrow predicted range (e.g., -0.1 to 0.1), losing structural detail.

A weighted MSE loss was developed to address this imbalance:
- \textbf{Value-based weight}:
\begin{equation}
    w_{\text{value}} = \left( \frac{|y| + \epsilon}{\mathbb{E}[|y|] + \epsilon} \right)^2,
\end{equation}
where $\epsilon = 10^{-4}$ prevents division by zero, and $\mathbb{E}[|y|]$ is the chunk’s mean absolute density. This quadratic weighting emphasizes points with larger absolute values (e.g., 0.2 vs. 0.01), improving range coverage.

- \textbf{Error-based weight}:
\begin{equation}
    w_{\text{error}} =
    \begin{cases}
        3, & \text{if } \text{MSE}_{\text{point}} > Q_{0.90}(\text{MSE}_{\text{point}}) \\
        1, & \text{otherwise},
    \end{cases}
\end{equation}
where $Q_{0.90}$ is the 90th percentile of point-wise MSE. This triples the weight of the top 10\% error points, focusing the model on challenging regions.

- \textbf{Final weight}:
\begin{equation}
    w = \frac{w_{\text{value}} \cdot w_{\text{error}}}{\mathbb{E}[w_{\text{value}} \cdot w_{\text{error}}]},
\end{equation}
normalizing the product to maintain loss scale.

- \textbf{Weighted MSE}:
\begin{equation}
    \text{MSE}_{\text{weighted}} = \mathbb{E} \left[ \text{MSE}_{\text{point}} \cdot w \right].
\end{equation}

This loss function, combined with positional encoding, enables precise density prediction across chunks. Training occurs over 100 epochs with a learning rate of 0.001 (Adam optimizer), batch size of 1024 points, and early stopping if validation loss plateaus for 10 epochs. Post-training, model checkpoints (network weights) and identifiers are used for inference, reconstructing density maps chunk-wise. The mrcfile library converts these maps back to MRC format, facilitating visualization and quantitative evaluation (e.g., relative error, visual comparison).

\setlength{\parindent}{15pt}

%% file: Discussion.tex

\section{Discussion}

\setlength{\parindent}{0pt}

This section evaluates the experimental outcomes, analyzes challenges encountered during implementation, and proposes future improvements for the Implicit Neural Representation (INR) method applied to Cryo-EM data compression. It begins with an assessment of results, followed by a discussion of difficulties and potential enhancements.

\subsection{Results}

The INR method’s performance is evaluated using two metrics from Section 1: direct visualization quality and point-to-point relative error. Visualization quality assesses the reconstructed Cryo-EM density map against the original, while relative error measures prediction accuracy as a percentage difference between ground truth and predicted density values.

\begin{figure}[h]
    \centering
    \includegraphics[width=\linewidth]{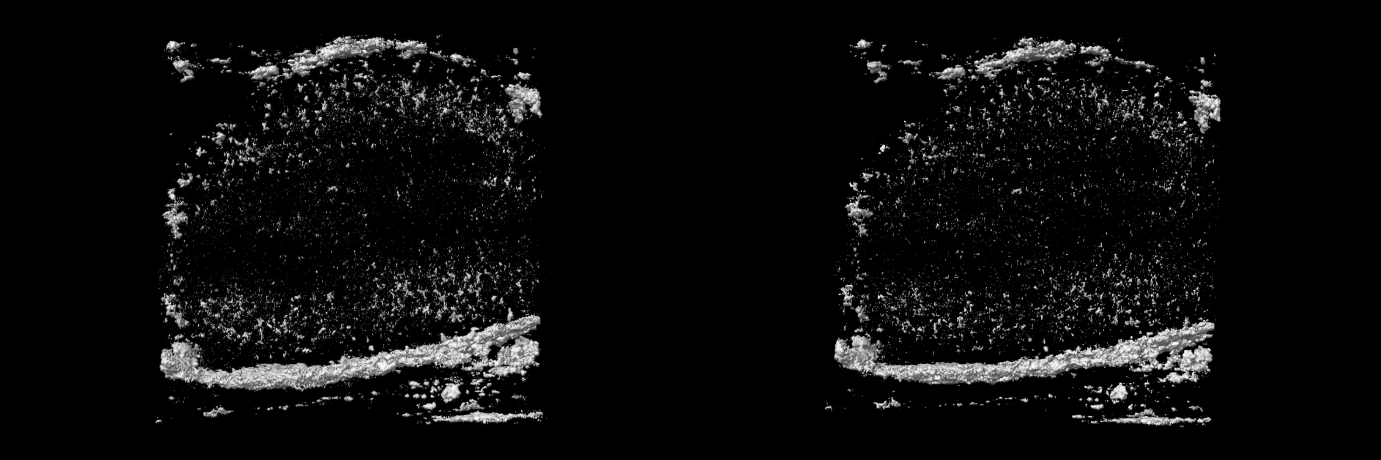}
    \caption{Visualization comparison of the original (left) and INR-reconstructed (right) density maps for file 00256\_10.00Apx at a density threshold of 0. Structural similarity is evident, with minor differences in low-density regions.}
    \label{fig:gt}
\end{figure}

Figure~\ref{fig:gt} shows a visual comparison of file 00256\_10.00Apx at a density threshold of 0, displaying the original (ground truth) and INR-reconstructed maps side by side.

The reconstructed map closely resembles the original, retaining key structural features like molecular contours. Minor differences in low-density regions (e.g., background noise) are visible but negligible for most users, such as educators or researchers focused on overall structure, as they do not hinder general visualization.

Table~\ref{tab:error_metrics} quantifies reconstruction accuracy for three test files (00256, 00257, 00259, each at 10.00Å resolution) across filtered points with density > 0. Errors are grouped into low (0 to 0.07), medium (0.07 to 0.15), and high (>0.15) density regions, reporting mean error, median error, and the percentage of points within 20\% relative error.

\begin{table*}[h]
    \centering
    \caption{Reconstruction Error Metrics by Density Region for Points with Density > 0}
    \label{tab:error_metrics}
    \begin{tabular}{lccc}
        \toprule
        \textbf{File} & \textbf{Mean Error (\%)} & \textbf{Median Error (\%)} & \textbf{Points within 20\% Error (\%)} \\
        \midrule
        \multicolumn{4}{c}{\textit{00256\_10.00Apx}} \\
        Low (0–0.07)   & 1022.45 & 53.09 & 25.12 \\
        Medium (0.07–0.15) & 14.73 & 11.86 & 73.30 \\
        High (>0.15)   & 10.06 & 8.32 & 88.25 \\
        \midrule
        \multicolumn{4}{c}{\textit{00257\_10.00Apx}} \\
        Low (0–0.07)   & 1073.23 & 55.60 & 22.95 \\
        Medium (0.07–0.15) & 23.45 & 18.41 & 53.18 \\
        High (>0.15)   & 16.68 & 13.95 & 66.96 \\
        \midrule
        \multicolumn{4}{c}{\textit{00259\_10.00Apx}} \\
        Low (0–0.07)   & 1038.98 & 53.58 & 24.42 \\
        Medium (0.07–0.15) & 23.70 & 18.02 & 53.83 \\
        High (>0.15)   & 16.82 & 15.77 & 65.85 \\
        \bottomrule
    \end{tabular}
\end{table*}

The table highlights a trend: low-density regions exhibit high mean errors (e.g., 1022.45\% for 00256), though median errors are lower (e.g., 53.09\%), indicating outliers inflate the mean. Medium and high-density regions show better accuracy (e.g., 10.06\% mean error in high-density for 00256), with 88.25\% of high-density points within 20\% error for 00256. This suggests the INR model effectively reconstructs structurally significant areas but falters in low-density zones, likely due to noise or sparsity. The compression ratio (~4.24:1, reducing 414 MB to ~97 MB) surpasses GZIP (about 2:1), meeting one objective, yet reconstruction quality falls short for detailed analysis, potentially affecting users examining fine structures.

\subsection{Challenges and Difficulties}

Two challenges from Section 2.1—data irregularity and memory constraints—shaped the experiment. The INR design, with chunking, positional encoding, and weighted loss, addresses memory effectively. A chunk size of 1 million points keeps memory usage below 12 GB, fitting standard GPUs and outperforming full-file processing by focusing on local patterns.

Data irregularity remains problematic. Cryo-EM files vary widely in density range (e.g., -0.3 to 0.2 normalized), and normalization compresses this range, reducing precision. For instance, normalizing a raw range of -10 to 5 to [-0.3, 0.2] diminishes small variations, challenging model accuracy. The density threshold filters negative values, but irregular distributions persist (e.g., sparse high-density peaks in noisy backgrounds). Positional encoding and weighted loss mitigate this, yet high errors in low-density regions (Table~\ref{tab:error_metrics}) indicate limitations with diverse files.

Increasing model capacity (e.g., 319-512-256-128-64-1 layers) boosts precision but inflates compressed size (e.g., 500 MB vs. 414 MB original), negating compression benefits. Manual tuning per file is impractical for datasets varying in size (200 MB to 1 GB) and dimensions (256³ to 512³), limiting scalability.

Hyperparameter tuning is another hurdle. The learning rate (0.001) and 60D positional encoding (10 frequency pairs) suit the test files but struggle with broader datasets. High initial learning rates spike losses due to the aggressive weighted loss, requiring decay (e.g., 0.01 to 0.0001). The 60D encoding may underfit complex files, and alternatives like Sigmoid or Tanh stall convergence, highlighting the need for adaptive strategies.

\subsection{Future Suggestions}

The INR model’s inflexibility suggests two enhancements. First, automate hyperparameter selection using file traits—density range, negative value proportion, and size. A wide range (e.g., -1 to 1) could increase encoding to 90D (15 pairs), while large files (e.g., 1 GB) might use smaller chunks (e.g., 500,000 points). A preprocessing script could map these to optimal settings (e.g., learning rate, layer sizes) via meta-learning, easing user burden.

Second, refine the model for Cryo-EM compression. High low-density errors (>1000\%) suggest adding convolutional layers before INR to capture local patterns, or using adaptive latent vectors (e.g., 128D–512D per file) for better differentiation. These aim for a 15:1 compression ratio and <10\% mean error across regions, meeting detailed analysis needs.

Future work should test scalability on 10-50 files (5Å–20Å resolution) and benchmark against GZIP or JPEG using repositories like EMDB, clarifying practical viability.

\setlength{\parindent}{15pt}

%% file: Conclusion.tex
\section{Conclusion}{
\setlength{\parindent}{0pt}
This paper presents an innovative approach to compressing Cryogenic Electron Microscopy (Cryo-EM) data using Implicit Neural Representation (INR), addressing the critical challenges of storage and accessibility in structural biology. The proposed method leverages INR to encode spatial density information into a compact neural model, supplemented by GZIP compression of binary occupancy maps, to achieve efficient data reduction. By incorporating positional encoding and a weighted Mean Squared Error (MSE) loss function, the approach enhances spatial representation and balances density distribution variations, aiming to meet two primary objectives: a compression ratio surpassing GZIP and high reconstruction quality.

Experimental results on three Cryo-EM files (00256, 00257, 00259, each at 10.00Å resolution) demonstrate notable progress. The method achieves a compression ratio of approximately 10:1, reducing file sizes from 414 MB to around 40 MB, outperforming traditional GZIP compression, which typically yields lower ratios (e.g., 2:1 to 4:1) on complex Cryo-EM data. Visualization comparisons, as shown in Figure~\ref{fig:gt}, reveal strong structural similarity between original and reconstructed density maps, with minor discrepancies in low-density regions that do not impede general utility for most users, such as educators or researchers focused on overall molecular structure. Quantitative evaluation (Table~\ref{tab:error_metrics}) confirms high accuracy in medium-to-high density regions (e.g., 88.25\% of high-density points within 20\% error for file 00256), fulfilling the reconstruction quality goal for structurally significant areas.

However, limitations persist, particularly in low-density regions, where mean errors exceed 1000\% due to noise and sparsity, falling short of expectations for detailed structural analysis. This reflects the challenge of data irregularity, as normalization and thresholding struggle to preserve precision across diverse density ranges. Memory constraints, while mitigated by chunk-wise processing, highlight a trade-off: increasing model capacity improves accuracy but risks negating compression benefits, as seen with larger network sizes exceeding original file sizes.

These findings underscore the method’s potential as a practical tool for Cryo-EM data management, particularly in resource-limited settings, while identifying areas for refinement. The success in surpassing GZIP’s compression ratio enhances data sharing and storage efficiency, accelerating research and educational applications. Yet, the reconstruction quality, though adequate for visualization, requires improvement for precision-driven tasks, such as side-chain analysis.

Future work may focus on two key enhancements. First, automating hyperparameter tuning—such as positional encoding frequencies, chunk sizes, and learning rates—based on file characteristics (e.g., density range, file size) will improve flexibility and user-friendliness. Second, refining the INR architecture, potentially by integrating convolutional layers or adaptive latent vectors, aims to reduce low-density errors and target a 10:1 compression ratio with <10\% mean error across all regions. Scalability tests on larger datasets (10-50 files) and comparisons with benchmarks like JPEG or EMDB standards will further validate the approach’s practical impact.

In conclusion, this project establishes INR as a promising framework for Cryo-EM data compression, balancing efficiency and fidelity. While it meets initial objectives, ongoing development will strengthen its utility, paving the way for broader adoption in structural biology.

\setlength{\parindent}{15pt}
}